%% file: ca.tex
\documentclass[10pt,twocolumn,letterpaper]{article}

\usepackage[pagenumbers]{cvpr} 

\input{preamble}

\definecolor{cvprblue}{rgb}{0.21,0.49,0.74}
\usepackage[pagebackref,breaklinks,colorlinks,citecolor=cvprblue]{hyperref}

\def\paperID{4949} 
\def\confName{CVPR}
\def\confYear{2024}

\title{Context-aware Talking Face Video Generation}

\author{Meidai Xuanyuan$^1$, Yuwang Wang$^2$, Honglei Guo$^2$, Qionghai Dai$^2$ \\
Tsinghua University\\
{\tt\small $^1$ xymd22@mails.tsinghua.edu.cn,}\\
{\tt\small$^2$\{wang-yuwang, guohonglei,  qhdai\}@mail.tsinghua.edu.cn,}
}

\begin{document}
\maketitle
\input{0_abstract}    
\input{1_intro}

\input{2_relatedwork}
\input{3_method}

\input{4_experiment}
\input{5_conclusion}

{
    \small
    \bibliographystyle{ieeenat_fullname}
    \bibliography{ca}
}

\input{6_supp}

\end{document}

%% file: preamble.tex
\usepackage[dvipsnames]{xcolor}


%% file: 0_abstract.tex
\begin{abstract}
In this paper, we consider a novel and practical case for talking face video generation. 
Specifically, we focus on the scenarios involving multi-people interactions, where the talking context, such as audience or surroundings, is present. 
In these situations, the video generation should take the context into consideration in order to generate video content naturally aligned with driving audios and spatially coherent to the context. 
To achieve this, we provide a two-stage and cross-modal controllable video generation pipeline, taking facial landmarks as an explicit and compact control signal to bridge the driving audio, talking context and generated videos. 
Inside this pipeline, we devise a 3D video diffusion model, allowing for efficient contort of both spatial conditions (landmarks and context video), as well as audio condition for temporally coherent generation. 
The experimental results verify the advantage of the proposed method over other baselines in terms of audio-video synchronization, video fidelity and frame consistency. 
\end{abstract}

%% file: 1_intro.tex
\section{Introduction}
\label{sec:intro}

\hspace{1em} Talking head video generation aims to create a harmonious and consistent video conditioned on a given audio~\cite{zhang2020davd,zhou2020makelttalk,zhou2021pose,shen2023difftalk,zhang2023sadtalker}, which has widely applications in the scenarios such as digital human creation and virtual avatars, etc. 
Most of the previous works focused on the talking head itself, without consideration of the audience and surroundings, referred as talking context. In some application scenarios, we may assume the audience talking to is in front or assigned, which is the typical case previous works focusing on\cite{wang2021oneshot,shen2022learning,shen2023difftalk,zhang2023sadtalker}. 
In some other application scenarios, for example, in the case of a video containing both talking person and the audience, it is important to consider the context when generating the talking head video, for example, the generated talking head should be oriented towards the audience, making the video more natural. 
It is necessary to consider the situations where the context content is present, such as inpainting head region in a video, interacting with the digital human, generating multi-person video, etc. 

\begin{figure}
    \centering
    \includegraphics[width=1 \columnwidth]{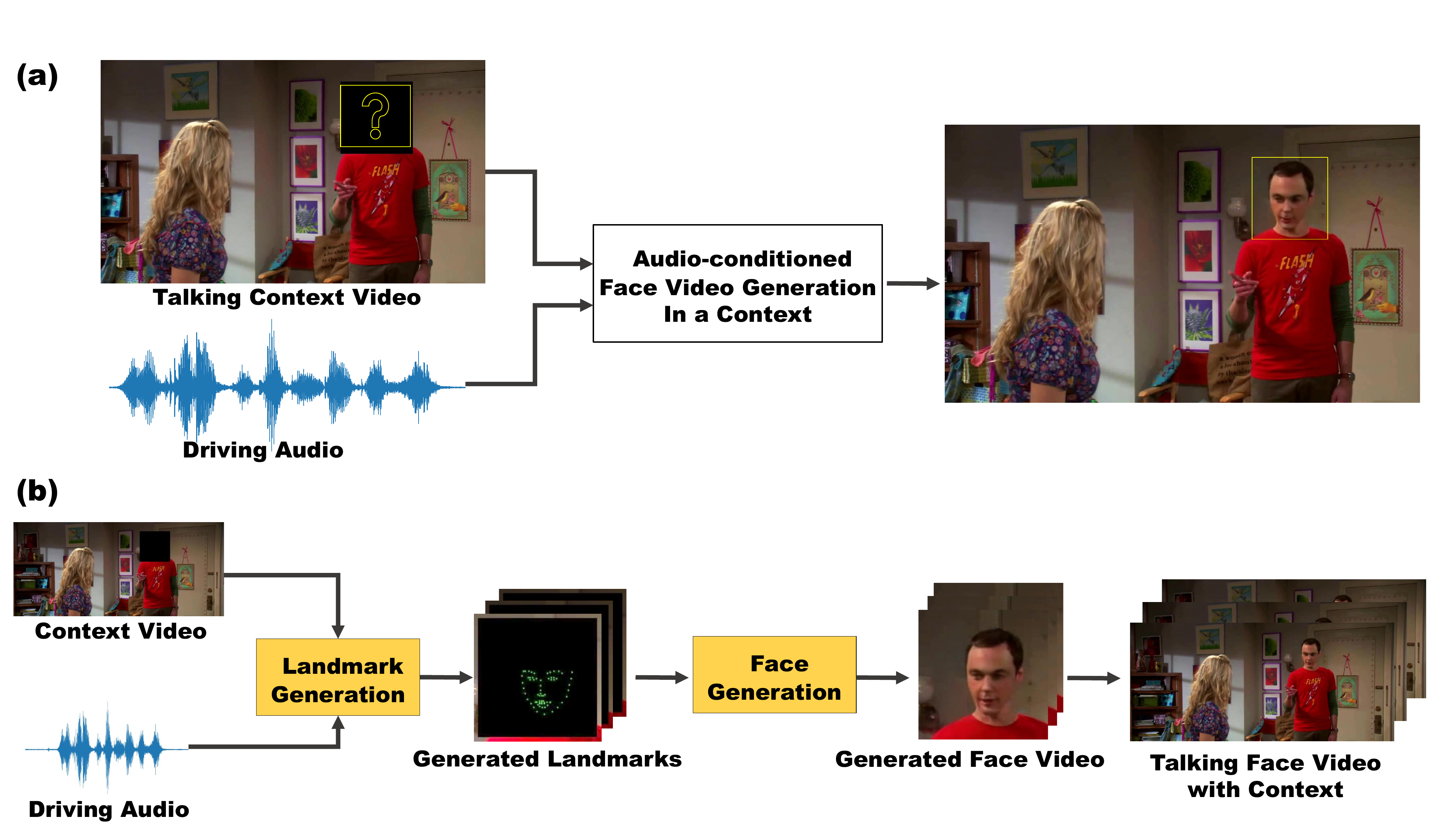}
    \caption{ (a) An illustration of our task setting. Given a context video and a driving audio, the task is to generate harmonious and consistent talking head video inside the masked region. (b) Our two-stage generation pipeline. Facial landmarks are adopted as an intermediate representation to enhance controllability. }
    \label{fig:task}
\end{figure}

Different from previous works, in this paper, we consider a novel and interesting setting for talking head video generation, i.e., generating conditioned on a context. Specifically, we propose a new setting as shown in Figure~\ref{fig:task}, given a driving audio and a video where the head region of target person is masked, serving as the context condition, the task is to generate a talking head video inside the masked region. From the perspective of context, this can also be regarded as video generation in the region of human head when the audio is given. Different from general video inpainting task, our setting focuses on talking face content and the generated video should be consistent with the driving audio.

To achieve audio conditioned talking face video generation given context video, we need to generate talking head video that: $(i)$ well aligned to the driving audio, not only in lip movements but also in facial expressions and head pose, and $(ii)$ well aligned to the conversation scene, which requires the model to understand scene semantics and generate coherent faces. The above two requirements are highly related to the landmarks of faces. 
With this observation, we devise a Two-stage Cross-modal Control generation pipeline, named TCCP, adopting facial landmarks as an explicit and compact intermediate representation to bridge the driving audio, video context and generated face videos. Specifically, in the first stage, the model focuses on understanding the context and generate a spatially compact representation (facial landmarks), transferring the implicit guidance from context into explicit facial landmarks. In the second stage, the model puts more efforts on the face video generation quality provided a video-level explicit key points guidance as condition.  
Compared to an end2end generation pipeline, our proposed TCCP explicitly models the key feature, facial landmark, making the generalization more controllable, promoting better temporal correlation with driving audio and spatial coherent with context video. 
Meanwhile, due to the compact property of facial landmarks, this design hold the potential for better generalization across identifies.
Inside TCCP, we propose a multimodal controlled video generation network, named MVControlNet,  which is a diffusion-based video generation model taking multi-type of conditions as input. Inspired by ControlNet~\cite{Zhang_2023_ICCV}, we design two branches inside MVControlNet, one is a 3D content diffusion model with conditions injected in the latent space\cite{ramesh2022hierarchical}, suitable for identity or audio conditions, i.e., semantics or cross-modal data. Another branch is a control branch receiving video-based conditions (such as landmarks, sharing the same shape with diffusion noise input) for finegrained spatial control.

In this paper, since there is no suitable existing dataset, we consider a practical talking scenario, videos from TV shows, which contain plentiful and diverse interaction between the talking person and context. We collected a dataset from \emph{The Big Bang Theory} containing 5.92 hours of talking videos of Sheldon, who is one of the main characters and has longest talking videos. 
In this paper, we focus on studying how to leverage the visual context and audio control signal to achieve visually natural and cross-modal consistent talking face videos. Due to the data and computing resource limitation, we narrow down to personal talking video generation for Sheldon. Our intermediate compact landmark representation is designed for generalization. ControlNet~\cite{Zhang_2023_ICCV} has demonstrated the effectiveness of using landmark-like spatial control single for image generation cross identities. 
Further, the generalization to other faces can be achieved by pretraining on a large-scale talking face dataset, then finetuning on a specific personal data, which has been verified in the previous face editing work~\cite{ding2023diffusionrig}. 

To verify the effectiveness of the proposed method, we conduct both objective and subjective experiments to evaluate the generated video in terms of audio-visual synchronization, video fidelity and frame consistency. Since we consider a new setting, we build several baselines based on some popular talking head methods and video generation techniques\cite{Fu_2022_CVPR}. Our proposed TCCP surpasses the baselines and achieves high quality audio conditioned talking face video generation in a context. 

Our main contributions can be summarised as follows:
\begin{itemize}
    \item We introduce an interesting, practical and novel setting for talking face video generation: taking the talking context into consideration.
    \item We devise a two-stage cross-modal control video generation pipeline to achieve audio conditioned talking face video generation in a context.
    \item We provide a MVControlNet to efficiently generate videos or facial landmarks with driving audio and context video conditions.
    \item We verify the effectiveness of the proposed method by comparing it with several baselines based on SOTA talking head or video generation techniques.
\end{itemize}

%% file: 2_relatedwork.tex
\section{Related Work}
\label{sec:relawork}

\hspace{1em}\textbf{Conditional Video Generation.}
Conditional video generation synthesizes a video guided by the given conditions, usually in the form of text description. Based on the remarkable results of text-to-image diffusion models, many works utilize pretrained T2I model and extend to video generation by finetuning\cite{singer2022makeavideo,zhou2023magicvideo,zhang2023controlvideo,qin2023dancing,ma2023follow,li2023videogen,hong2022cogvideo}. Multiple techniques for improving time consistency are suggested, including adding in temporal attention layers\cite{zhang2023controlvideo} or 3D convolution blocks\cite{singer2022makeavideo,zhou2023magicvideo}. By adding in ControlNet structure, video generation can be also guided by depth maps or pose sequences\cite{chen2023controlavideo,zhang2023controlvideo,qin2023dancing,ma2023follow}. Some works bridge the diffusion process between modalities and can synthesize videos according to audio clips\cite{Ruan_2023_CVPR,pmlr-v202-bao23a}. However, compared to our explicit facial landmarks,  the conditions are relatively less strict and the generated contents are less controllable.

\textbf{Audio-driven Talking Head Synthesis.}
Despite the works\cite{Ruan_2023_CVPR,pmlr-v202-bao23a} mentioned above, talking head synthesis also aims to generate talking videos with lip movements synchronized with the audio condition. The existing methods can be roughly divided into two groups: 2D-based methods and 3D-based methods. 2D-based methods usually utilize GANs\cite{doukas2021headgan,das2020speech,Prajwal_2020_wav2lip,zhou2020makelttalk} to learn lip-audio synchronization. However, due to the instability of GANS, the generated videos tend to be of limited image quality as well as relatively low resolution\cite{dhariwal2021advances}. More recently, 3D-based methods\cite{song2022everbody,thies2020neural,liu2022semanticaware,shen2022learning} employ 3D Morphable Models\cite{blanz2023a} or Neural radiance fields\cite{mildenhall2020nerf} for parameterized modeling. While greatly improving visual quality, most of these methods are difficult to generalize across different identities. Moreover, works dealing with talking head synthesis mainly focus on the quality of generated lip motions, while lacking head pose diversity and facial details like eye movements or facial expressions. The impact of the conversation context, is also out of consideration.

\textbf{Video-to-Video Synthesis.}
Like conditional video generation, many video-to-video translation models are built on the basis of image synthesis architecture. Fu et al.\cite{Fu_2022_CVPR} introduces the language-based video editing task which translates from video to video following textual instructions. In this field, Gen-1\cite{esser2023structure} and FateZero\cite{QI_2023_ICCV} utilize pretrained text-to-image diffusion model and edit the target object continuously by adding temporal attention. Video-P2P\cite{liu2023videop2p} and VideoControlNet\cite{hu2023videocontrolnet} choose to perform frame-by-frame translation, then reform the video through cross frame attention or temporal interpolation. StableVideo\cite{Chai_2023_ICCV} employs pretrained NLA model\cite{lu2020} for text-guided appearance transfer, and keeps appearance consistency by introducing temporal dependency constraints. Restricted by relatively high dimension, most existing models mainly focus on editing the whole video scene or major object generally, lacking support for fine-grained video manipulation. Also, the translation is mainly guided by given conditions rather than an understanding of video semantic. 

\textbf{Conditional Image Manipulation.}
Some previous works utilized generative adversarial networks (GANs) to perform such image manipulations\cite{NEURIPS2020_6fe43269,lang2021,abdal2020,alaluf2021,omer2021,ni2023}. Other deep techniques have also been applied to image editing\cite{esser2021,bartal2022,yu2022}, and can follow multi-modal conditions like pose skeletons or canny edges.
More recently, with the improvement in diffusion-based text-to-image synthesis quality, pretrained diffusion models have achieved remarkable results. 
Given background mask and text prompt, Imagen Editor\cite{su2023} and Blended Diffusion\cite{Avrahami_2022_CVPR} can modify the masked area without affecting the unmasked regions, on the basis of CLIP-guided diffusion for text semantic understanding. 
Some works\cite{xie2023} make use of multi-modal conditions other than text, and ControlNet\cite{Zhang_2023_ICCV} introduces an architecture for adding in self-designed conditional control. There are also works that discard text prompts entirely and repaint only guided by background mask\cite{Lugmayr_2022_CVPR}.
Although the conditional image editing techniques mentioned above can perform detailed image manipulation, they heavily rely on the given conditions for image understanding and task comprehension. Also, constrained by modality, these models require extra modifications to input time-continuous conditions like audio or pose sequence. 

%% file: 3_method.tex
\section{Preliminary: Latent Diffusion \& ControlNet}
\begin{figure*}
    \centering
    \includegraphics[width=1 \linewidth]{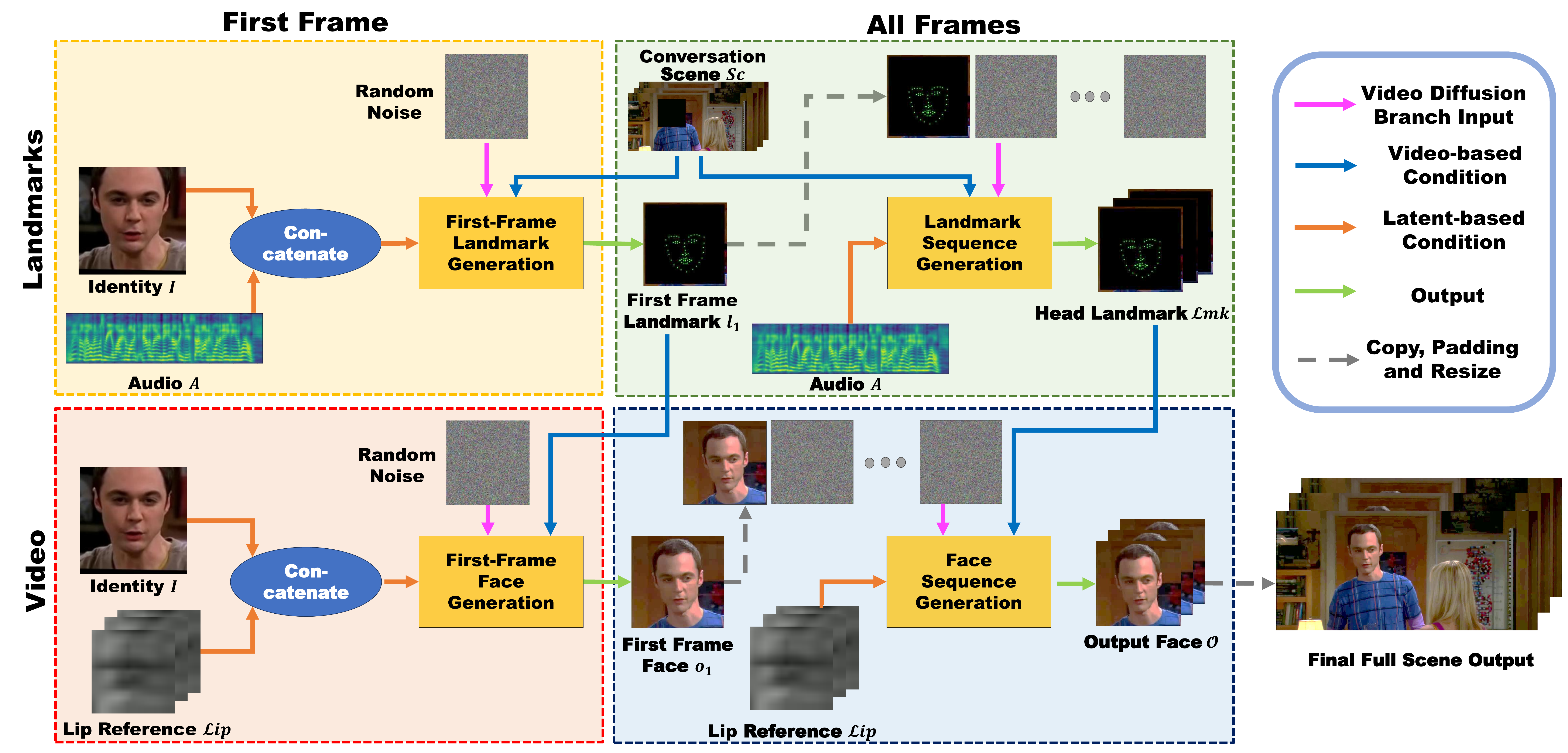}
    \caption{TCCP pipeline. \textbf{Rows referring to pipeline stages}: taking head facial landmark $\mathcal{L}mk$ generation and coherent talking face video $\mathcal{O}$ generation. \textbf{Columns referring to temporal steps}: First Frame and All Frames. Each step is built on MVControlNet, a two-branch diffusion-based model. The video diffusion branch takes in Video Diffusion Branch Input and Latent Based Condition. The control branch takes in Video-based Condition. Different types of inputs/outputs are marked by arrows in different colors.}
    \label{fig:task_pipeline}
\end{figure*}
\hspace{1em}The forward process of Denoising Diffusion Probabilistic models (DDPMs)\cite{ho2020denoising} is performed over a discrete $T$ time step. Given $x_0$ as an input sample from input distribution $p(x)$, and $x_T$ is sampled from a unit Gaussian independent from $x_0$ utilizing the Markovian forward process, then the diffusion forward process can be defines as:
\begin{equation}
    q(x_t|x_{t-1})=\mathcal{N}(x_t;\sqrt{1-\beta_t}x_{t-1},\beta_tI).
\label{Eq:q_sample}
\end{equation}
\begin{equation}
    q(x_{1:T}|x_0)=\prod_{t=1}^Tq(x_t|x_{t-1}).
\label{Eq:q_sample_chain}
\end{equation}
Here $t\in[1,T]$, and $\beta_0,\beta_1,\dots,\beta_T$ are predefined variance schedule sequences. Following previous works\cite{ho2020denoising,song2021scorebased}, we increase $\beta_t$ using linear noise schedule. The forward process depicts a gradual diffusion schedule in which a noise depending on variance $\beta_t$ is added to $x_{t-1}$ to get $x_t$ every time step and finally reaches $x_T$. The goal of diffusion model is to reverse diffusion process and recover $x_0$ from $x_T$, such reverse process can be formulated as:
\begin{equation}
    p(x_{t-1}|x_t)=\mathcal{N}(x_{t-1};\mu_{\theta}(x_t,t),\Sigma_{\theta}(x_t,t)).
\label{Eq:p_sample}
\end{equation}
\begin{equation}
    p_{\theta}(x_{0:T})=p(x_T)\prod_{t=1}^Tp(x_{t-1}|x_t).
\label{Eq:p_sample_chain}
\end{equation}

Here $\mu_{\theta}$ refers to the Gaussian mean value predicted by model $\theta$, $x_T$ is a given random noise, and the variance $\Sigma_{\theta}(x_t,t)$ is fixed in our experiments as it only leads to minor improvement\cite{bao2022analyticdpm}. The model can be then denoted as a denoising model $\epsilon_{\theta}(x_t,t)$, and is trained to predict the noise of $x_t$. The noise prediction loss can be defined as:
\begin{equation}
    min_{\theta}||\epsilon-\epsilon_{\theta}(x_t,t,c_l)||_2^2,
\label{Eq:loss}
\end{equation}
where $\epsilon$ is the actual noise added to $x_t$, and $c_l$ refers to latent conditions like audio or lip reference. So the model learns to predict the noise of $x_t$ conditioned on $c_l$ at time step $t$.

Due to the relatively high dimension of videos, we employ an auto-encoder $\mathcal{E}$ to compress input $x$ into latent code $z$ following Latent Diffusion\cite{ramesh2022hierarchical}. The model then learns to denoise in latent space, and reconstructs latent code $z_0$ during inference. A pretrained decoder $\mathcal{D}$ is utilized to obtain the generated output as $x_0=\mathcal{D}(z_0)$.

To learn from additional task-specific semantic conditions like masked scene or facial landmark, we follow ControlNet\cite{Zhang_2023_ICCV} to add a control branch to diffusion model. The architecture of ControlNet enables finetuning generative diffusion model on additional conditions quickly by utilizing trainable layers copied from the original model. This added condition should be of the same shape with $x_t$. In our video generation task, given the additional video semantic condition $c_v$, the noise prediction loss can be re-defined as:
\begin{equation}
    min_{\theta}||\epsilon-\epsilon_{\theta}(x_t,t,c_l,c_v)||_2^2.
\label{Eq:control_loss}
\end{equation}
In our video-to-video synthesis task, both the input $x$ and condition $c,c_s$ are in the form of N-frame sequences. Thus, additional temporal layers are added to ensure frame consistency and temporal in-context comprehension.

\section{Contextual Talking Face Video Generation}
\label{sec:method}

\subsection{Task Formulation}

\hspace{1em}We study the audio conditioned in-context talking head generation task, given a driving audio $\mathcal{A}$ of speaker $\mathcal{I}$, a context video $\mathcal{S}c$ where the head region of talking person is masked, the target is to generate a target video $\mathcal{O}$, which fills the talking person head region. The main objective of our task is that the generated video $\mathcal{O}$ should be: $(i)$ well aligned to the driving audio $\mathcal{A}$, not only in lip movements but also in facial expressions and head poses, and $(ii)$ spatially reasonable to the context $\mathcal{S}c$, such as oriented to the audience, coherent with gesture, etc. 
The $\mathcal{S}c$ contains $N$ frames, denoted as $\{s_{c1},\dots,s_{cN}\}$. The generated $\mathcal{O}$ should also contain $N$ frames $\{o_1,\dots,o_N\}$. The driven audio $\mathcal{A}$ is first transformed into mel spectrograms, resulting in acoustic features $\mathcal{A}=\{a_1,\dots,a_L\}$, where $L$ is the length of acoustic feature. Identity $\mathcal{I}$ can be a randomly selected picture depicting the target identity. 


\begin{figure*}
    \centering
    \includegraphics[width=1 \linewidth]{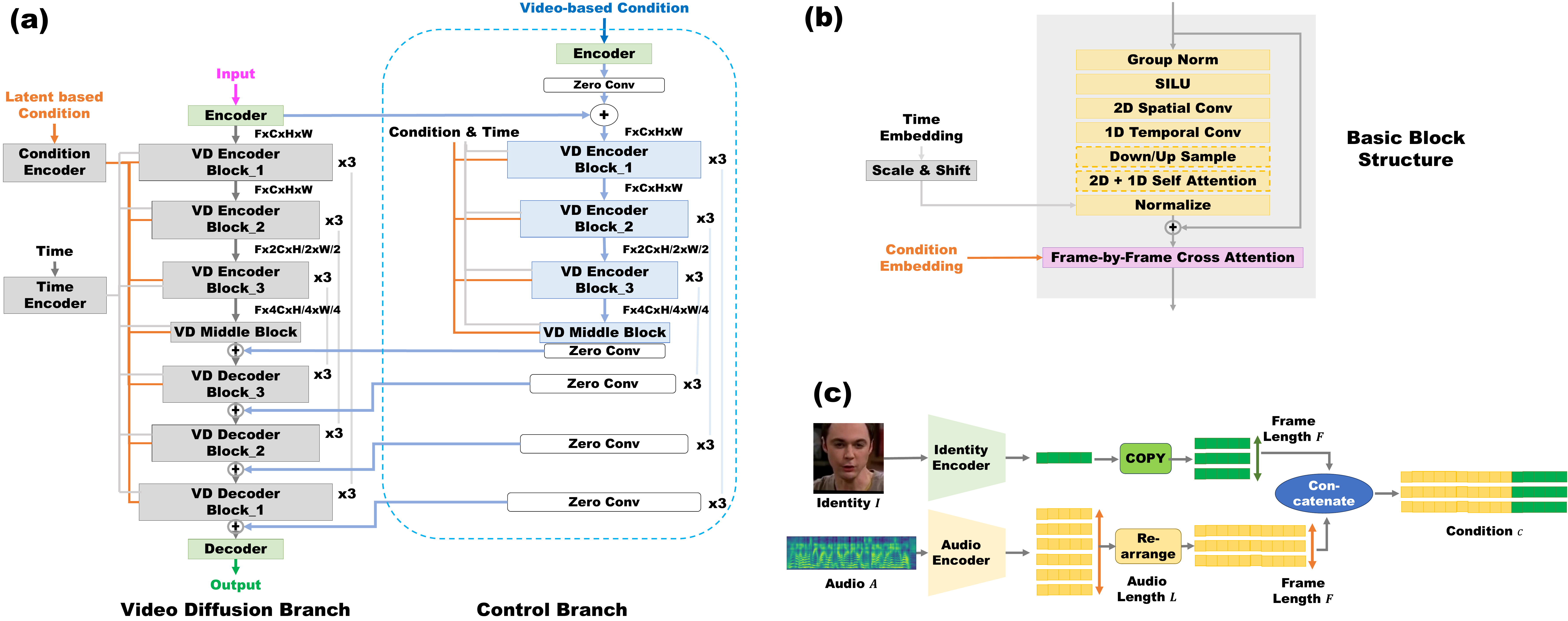}
    \caption{Model Architecture of multimodal controlled video generation network (MVControlNet). (a) The overall model architecture, consisting of two model branches. The Video Diffusion Branch predicts the diffusion noise added to the input at time step $t$ under the latent-based condition $c_l$. The Control Branch is a trainable copy of the video diffusion branch, which is used to add video-based condition $c_v$ to the video diffusion branch by zero convolution layer. (b) A detailed demonstration of Video Diffusion Block. 1D temporal convolutions and temporal self attention are stacked after spatial layers to enforce spatial-temporal coherence. (c) A brief illustration of encoding of latent-based condition $c_l$. Conditions of multiple modalities will be encoded separately, and then concatenated by frame. }
    \label{fig:model}
\end{figure*}

\subsection{TCCP: Two-stage Cross-modal Control generation Pipeline}

\hspace{1em}As mentioned before, we utilize the facial landmarks as an explicit and compact intermediate representation, denoted as $\mathcal{L}mk=\{l_1,\dots,l_N\}$ for $N$ frames, which has the same shape of $\mathcal{O}$. 
The main generation pipeline is shown in Figure.~\ref{fig:task_pipeline}.
We divide the generation pipeline into two stages: taking head facial landmark $\mathcal{L}mk$ generation and coherent talking face video $\mathcal{O}$ generation, shown in the first row and second row in Figure.~\ref{fig:task_pipeline}, respectively. 
In a temporal dimension, the pipeline can be divided into two step, First Frame and All Frames shown in Figure.~\ref{fig:task_pipeline} as first column and second column, respectively. Since we do not assume the first frame is provided, the First Frame case generates the first image and the related landmarks, which can be regarded as a special case of All Frames. 

In each step, at the heart, there is the proposed MVControlNet, which takes three inputs: Video Diffusion Branch Input, Video-based Condition and Latent Based Condition. The four MVControlNet can be regarded as different variants according to the content difference in terms of the input and output. Our MVControlNet is a two-branch diffusion-based model. The first branch is a video diffusion branch, generating video contents, such as the landmarks $\mathcal{L}mk$ or the generated video $\mathcal{O}$. This branch takes audios and identity as condition, which is injected in the latent space, denoted as latent-based condition $c_l$. The second branch is a control branch, which takes video-based control signal as input, which is injected to the video diffusion branch in ControlNet-like way, denoted as video-based condition $c_v$. 


\textbf{Talking head landmark generation} refers to the upper two blocks in Figure~\ref{fig:task_pipeline} and can be further divided into two temporal steps: first-frame landmark generation and landmark sequence generation. The target of this stage is to generate facial landmarks well aligned with driven audio and context video. Driven by this main purpose, this stage takes in head-masked conversation scene video $\mathcal{S}c$ as video-based condition $c_v$, and takes driving audio $\mathcal{A}$ as the latent-based condition $c_l$. The first step of the stage (first-frame landmark generation) additionally takes identity $\mathcal{I}$ as condition to guide facial shape generation. 
Then the output of the first step (the generated frame landmark $l_1$) will be utilized as the first frame condition for the next step. 
The second step of the stage (landmark sequence generation) generates several frames of coherent landmarks conditioned on given first frame. 
The final model output of this stage is the diffusion noise $\epsilon_{\theta}(x_t,t,c_l,c_v)$ added to a landmark sequence, and $\mathcal{L}mk$ can be obtained through DDPM. 

\textbf{Coherent talking face generation} refers to the bottom blocks in Figure~\ref{fig:task_pipeline}. The procedure of face generation stage is the same as the former landmark generation stage, in which the initial face frame will be generated first, followed by subsequent face sequence generation. This stage fills in face details like eye movements, facial expressions and appearances, ensuring that all these generated features are coherent with conversation scene context. Talking head landmark sequence $\mathcal{L}mk$ is utilized as video-based condition $c_v$, and lip sequence $\mathcal{L}ip$ as well as identity $\mathcal{I}$ are employed as latent-based conditions $c_l$ to guide lip/face shape generation. 
Similarly, the first frame face $o_1$ is generated in the first step and used as initial frame condition in the following face sequence generation step. The final output of this stage is the diffusion noise $\epsilon_{\theta}(x_t,t,c_l,c_v)$ added to a face sequence, and through DDPM a face video $\mathcal{O}$ can be obtained. With simple resizing and relocating, the generated face video can be filled into the original conversation scene, giving a complete, reasonable and coherent talking video.

\subsection{MVControlNet}


\hspace{1em}\textbf{The Video Diffusion Branch} uses U-Net as model structure, taking in encoded noise sequences $x_t$ as input and outputs predicted noise $\epsilon_{\theta}(x_t,t,c_l,c_v)$, under additional conditions $c_l$ and $c_v$. 
VD Encoder/Decoder/Middle Block is composed of multiple basic blocks, which is a spatial-temporal resnet block and is depicted in details in Figure~\ref{fig:model}(b) . 
There are two main architecture modifications compared to image diffusion models: $(i)$ 1D temporal convolutions are stacked after 2D spatial convolutions. We decompose the spatial and temporal dimensions following Ho et al.\cite{ho2022video} to avoid using heavy 3D convolutions while still adding temporal comprehension. $(ii)$ Similarly, temporal self attention is blocked after spatial self attention, to enforce spatial-temporal coherence.

Furthermore, the latent-based condition $c_l$ here is not restricted to a single form but can be a combination of acoustic feature, facial feature, lip reference or identity reference. 
As shown in Figure.~\ref{fig:model}(c), all these multimodal conditions will be encoded and concatenated frame by frame, which means the final condition code $c_l$ and input $x_t$ are of the same length in temporal dimension. 
Then cross attention between condition $c_l=\{c_{l1},\dots,c_{lN}\}$ and input $x_t=\{x_{t1},\dots,x_{tN}\}$ can be calculated frame by frame. The cross attention of frame $i$ can be formulated as:
\begin{equation}
CrossAtt(x_{ti},c_{li})=Softmax(\frac{Q_i^x(K_i^c)^T}{\sqrt{d}})V_i^c.
\label{Eq:cross_attn}
\end{equation}
where $Q_i^x = W_Qf(x_{ti})$, $K_i^c = W_Kf(c_{li})$, $V_i^c = W_Vf(c_{li})$,
and $f(\cdot)$ refers to flatten operation to 1 dimension. The cross attention is performed frame by frame to enforce understanding of corresponding temporal relation between input and latent-based conditions. Then the output is rearranged and concatenated in temporal dimension.

\textbf{The Control Branch} is a copy of the video diffusion branch, sharing identical model architecture. The control branch takes video-based condition $c_v$ as input, and the encoded $c_v$ will be added to the encoded noise input $x_t$ by zero convolution layer. 
The output of the control branch will be passed back to corresponding layer of the video diffusion branch through zero convolution layer. 
To speed up training process, the video diffusion branch can be reused, while the control branch has to be finetuned if given different types of video-based conditions.

\subsection{Training and Inference with First-frame Conditioning}

\hspace{1em}Following previous video generation works~\cite{chen2023controlavideo}, we utilize first-frame conditioning technique instead of generating the whole sequences to enable efficient training and recurrent inference of longer videos.  
This frees model from memorizing video content in the training set, so that the model can focus on reconstructing motion rather than the full frame. 
As depicted in Figure~\ref{fig:task_pipeline} All Frames column, we do not add noise to the first-frame input $v_1$ (which can also be specifically denoted as landmark $l_1$ or face $o_1$ in different task stages), so the model learns to generate subsequent frames based on $x_1$, guided by other conditions $c_l$, $c_v$ and time step $t$, and the loss function can be formulated as:
\begin{equation}
    min_{\theta}||\epsilon-\epsilon_{\theta}(x_t,t,c_l,c_v,\mathcal{E}(v_1))||_2^2.
\label{Eq:first_frame}
\end{equation}
Here $\mathcal{\epsilon}(v_1)$ refers to the latent code of first-frame input $v_1$. Utilizing the first-frame conditioning strategy, our model can make effective use of content from the first frame and better understand the motion guidance from the conditions. Furthermore, during inference, by conditioning on previously generated frames as the initial frame in subsequent iterations, our model can generate longer videos in an auto-regressive way.

Specifically, we separate first frame generation and subsequent sequence generation into two steps. During training, the model learns to first generate initial frame $v_1$, and this generated initial frame is used for subsequent sequence generation training auto-regressively. The model architecture of these two steps is similar. During inference, we generate the initial frame $v_1$ given Gaussian noise in the form of a single frame $x_1$, then following frames are generated conditioned on $\mathcal{E}(v_1)$ as described in Eq~\ref{Eq:first_frame}.

%% file: 4_experiment.tex
\section{Experiments}
\label{sec:exp}

\subsection{Experimental Setup}

\hspace{1em}\textbf{Dataset.}
We base our experiments on talking videos collected from a TV series \emph{The BigBang Theory}, for diverse conversation context and sufficient personalized data. Face detector\cite{Liu2016ssd} is applied to detect and crop out talking faces, then facial landmarks are extracted\cite{bulat2017far} as groundtruth of the first stage and video-based conditions of the second stage. Corresponding audios can be directly obtained from the video clips and used to drive pre-trained Wav2Lip\cite{Prajwal_2020_wav2lip} for lip references. As we focus on personalized model, we only collect data for the main character \emph{Sheldon}. The final dataset consists of about 66k data sequences, 8 frames long each.

\textbf{Evaluation Metric.} We adopt the following metrics:
\begin{itemize}
    \setlength{\leftskip}{0.1em}
    \item  {SyncNet confidence score\cite{Prajwal_2020_wav2lip}}: The SyncNet score evaluates lip synchronization and mouth shape quality, and can partly check audio-visual synchronization quality.
    \item  {FID\cite{heusel2018gans,Seitzer2020FID}}: Frechet Inception Distance is employed to evaluate visual fidelity. We calculate this metric on both generated faces and on the whole conversation scene, to check model's comprehension on overall talking context.
    \item  {Frame Consistency}: Following previous works\cite{esser2023structure,QI_2023_ICCV}, we compute CLIP image embeddings on all frames of output videos and report the average cosine similarity between consecutive frames. This metric measures semantic coherence.
\end{itemize}

\textbf{Baselines.}
Since we are introducing a brand new task, there is no existing baselines. We consider approximating our task using two types of works: $(i)$ Scene context guided image generation, which can be trained to understand spatial context conditions but lack timing mechanism; $(ii)$ Talking head synthesis, which can generate talking videos with lip movements synchronized with the driving audio. We adopt the following baselines:
\begin{itemize}
    \setlength{\leftskip}{0.1em}
    \item {ControlNet}\cite{Zhang_2023_ICCV}: ControlNet introduces an architecture for adding in self-designed conditional control into a pre-trained diffusion model. We employ this network architecture and finetune the pre-trained Stable Diffusion Model v1.5\cite{Rombach_2022_CVPR} to generate face images guided by given conversation scene and facial landmarks. The model is trained for 230k steps under default parameter setting. And the training data is kept same with that used in our TCCP, only split from sequences to single images due to lack of timing mechanism in original Stable Diffusion Model. As ControlNet can't be driven by audio, during inference, we directly employ groundtruth face landmarks as video-based conditions and generate frame by frame following $M^3L$\cite{Fu_2022_CVPR}.
    \item {SadTalker}\cite{zhang2023sadtalker}: SadTalker is a talking head generation model that creates natural lip motions well synchronized with driven audio. We run the model with deriven audio and groundtruth initial face frame.
\end{itemize}

\textbf{Implementation Details.}
For the Video Diffusion U-Net, we set 3 scales of Video Diffusion Blocks, and each is stacked by 2 basic blocks and 1 down/up-sample block. Self attention and frame-by-frame cross attention are applied on all U-net scales. For all experiments, we follow previous works\cite{ho2020denoising,song2021scorebased} to use a linear noise schedule and the diffusion step number $T$ is set to 1000. The input/output sequence length $\mathcal{N}$ is set as 8. And the downsampling factor for Latent Diffusion auto-encoder $\mathcal{E}$ is set as 16. Adam is adopted to optimize through our model with learning rate 2e-4. All experiments are carried out on 8 NVIDIA 3090Ti GPUs. As for audios, they are processed to 16kHz and transformed to mel spectrograms with the same setting as Wav2lip\cite{Prajwal_2020_wav2lip}. Details of model architecture and training configuration can refer to supplementary materials.

\subsection{Quantitative Results}

\begin{table}[t]
  \centering
  \resizebox{1\columnwidth}{!}{%
  \begin{tabular}{ccccc}
  \toprule[2pt]
    Model & {Sync. score $\uparrow$} & FID(Face) $\downarrow$ & FID(FullScene) $\downarrow$ & F.C. $\uparrow$ \\
    \hline
    GT Samples & 3.11 & 28.65 & 22.13 & 0.989 \\
    ControlNet & 1.89 & 131.28 & 45.60 & 0.979 \\
    SadTalker & 2.95 & 75.16 & 36.63 & \textbf{0.996} \\
    Ours(gt lmk) & \textbf{2.99} & \textbf{68.05} & \textbf{26.42} & 0.988 \\
    Ours(gen lmk) & 2.76 & 72.26 & 27.06 & 0.986 \\
  \specialrule{0em}{1pt}{1.5pt}
  \bottomrule[2pt]
  \end{tabular}%
  }
  \caption{Comparison of quantitative results with baselines. F.C. denotes Frame Consistency. The best performed other than groundtruth is highlighted in bold.}
  \label{tab:quantitative}
  \end{table}

\hspace{1em}Table~\ref{tab:quantitative} demonstrates the quantitative testing results compared between the baselines and our MVControlNet. All models are tested for same talking scenarios, and we employ our model under two strategies. \textbf{Ours(gt lmk)} refers to inference of the second stage (only Face Video Generation Stage) given groundtruth facial landmarks, totally aligned with ControlNet experimental settings. \textbf{Ours(gen lmk)} refers to inference of the full pipeline, starting from First Frame facial landmark generation, requiring the model to read the scene context and synchronize with driving audio. 

Even though ControlNet is powerful in image-to-image translation, the deficiency in temporal context understanding severely impacts output video quality. SadTalker, on the contrary, displays superior audio-visual synchronization ability but still not performs well in FID score. It is more obvious when considering FID (full scene), as the gap between SadTalker and our model is enlarged. This demonstrates that the absence of scene comprehension in model has a negative impact on generated video quality, especially in the context of full talking scenario. As for Frame Consistency, this metric evaluates perceptual consistency more over pixel-level visual consistency, thus all models score high on this metric. Since the head pose generated by SadTalker lacks diversity compared with groundtruth and our model, it ranks highest in Frame Consistency as all frames demonstrate high similarity.

\textbf{Our model performs well on all these metrics, and achieves high quality audio conditioned talking face video generation in a context.} Model guided by groundtruth landmarks performs best considering audio-visual synchronization and visual quality, demonstrating outstanding ability in utilizing multimodal conditions and understanding spatial-temporal context. Even under more difficult setting, i.e., the videos driven by generated landmarks still achieve relatively high score in all metrics, and outperform baselines in scene context comprehension.

\subsection{Qualitative Results}

\begin{figure}
    \centering
    \includegraphics[width=1 \columnwidth]{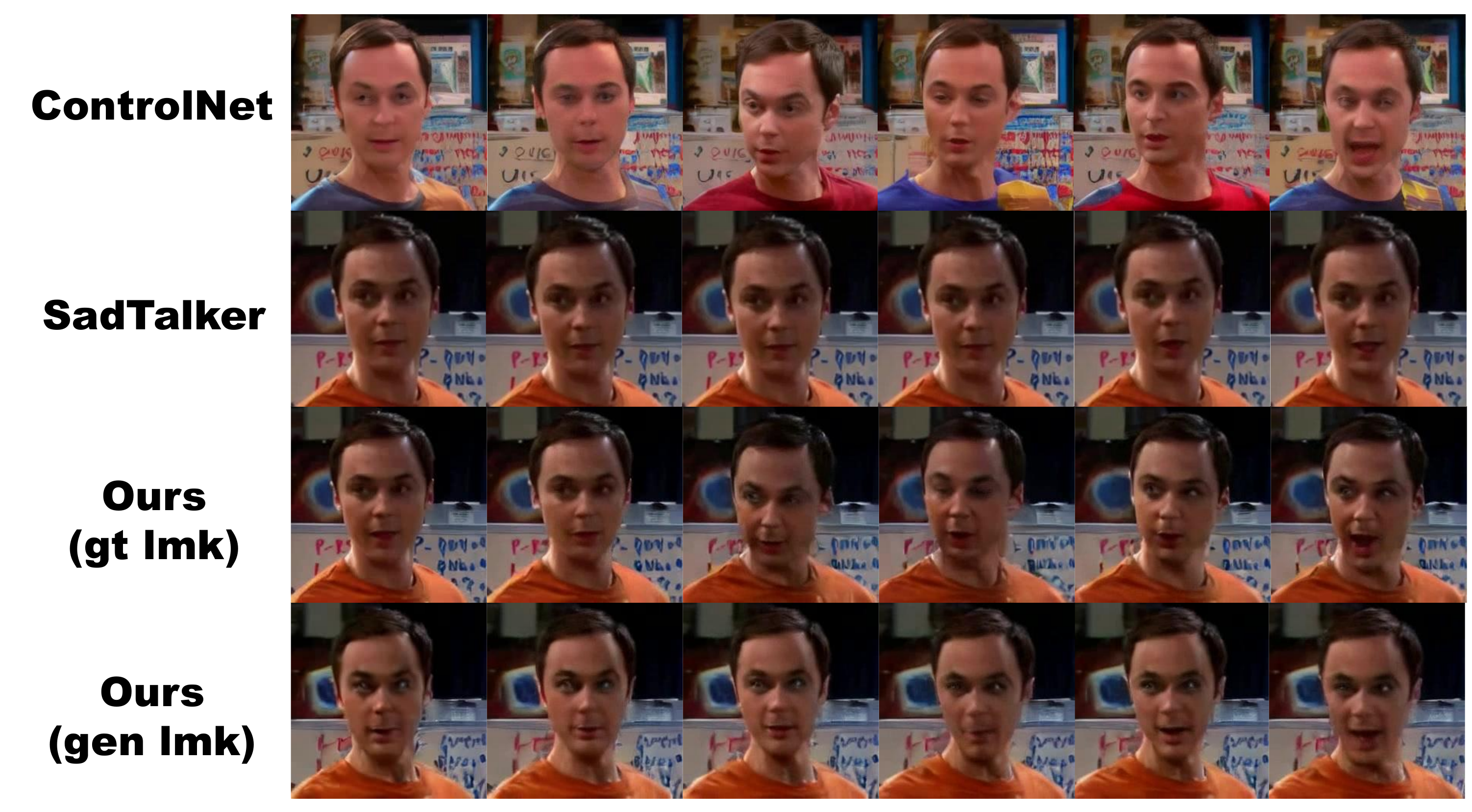}
    \caption{We compare our work with baselines in face generation quality. Our model can generate harmonious and consistent talking face with vivid facial expressions and diverse head poses.}
    \label{fig:face_quality}
\end{figure}

\begin{figure*}
    \centering
    \includegraphics[width=1 \linewidth]{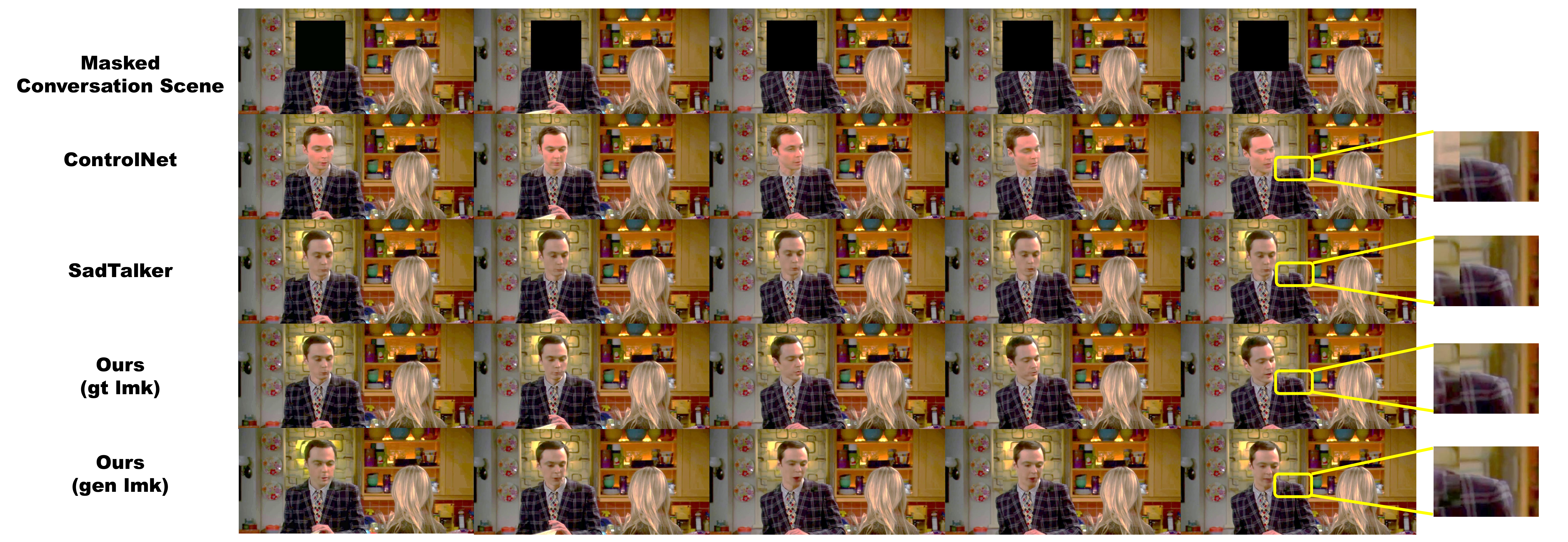}
    \caption{We compare our work with baselines considering full conversation scene quality. Scene details are given in the right column, and inconsistency can be observed near the shoulder in results generated by ControlNet and SadTalker.}
    \label{fig:full scene}
\end{figure*}

\hspace{1em}\textbf{Our model can generate harmonious and consistent talking face with vivid facial expressions and diverse head poses.} Figure~\ref{fig:face_quality} depicts the generated face videos compared with baselines. As ControlNet can only perform image-to-image translation, the resulting faces are incoherent in temporal dimension and keeps changing in character's clothes and hairstyle. SadTalker, however, gives fairly coherent results but lacks diversity. The generated heads stay relatively still and the character's eyesight and facial expressions seem unchanged. On the contrary, our model, demonstrates greater diversity in providing various head poses, facial expressions and vivid eye movements, while still maintaining high temporal consistency.

\textbf{Our model has stronger understanding of talking context and better fits in the conversation scene.} We also give examples of the final conversation scene outputs to check model's ability in fitting back to the full scenarios. As shown in Figure~\ref{fig:full scene}, ControlNet and SadTalker fail in giving spatial coherent results. An obvious inconsistency can be observed near the shoulder, as depicted in the right column of Figure~\ref{fig:full scene}. Contrarily, our model, no matter using groundtruth landmarks or generated landmarks, outperforms in generating faces that better fit the full scene.

\subsection{User Study}

\begin{table}[t]
  \centering
  \resizebox{1\columnwidth}{!}{%
  \begin{tabular}{ccc}
  \toprule[2pt]
    Model & Naturalness(Face) & Naturalness(Full Scene) \\
    \hline
    ControlNet & 2.07 & 2.31 \\
    SadTalker & 3.64 & 2.92 \\
    Ours & 3.62 & 3.94 \\
  \specialrule{0em}{1pt}{1.5pt}
  \bottomrule[2pt]
  \end{tabular}%
  }
  \caption{Results of user study.}
  \label{tab:user_study}
  \end{table}

\hspace{1em}32 participants were surveyed to evaluate the naturalness of the generated videos using a rating scale ranging from 1 to 5. Specifically, they were shown the generated face videos first and asked to score the naturalness of face alone. After face scoring finished, the full conversation scenarios were given , and the participants were asked to judge the naturalness once again depending on the overall video quality. The results are shown in Table~\ref{tab:user_study}, and are well aligned with quantitative results. In terms of face quality, our model obtains almost the same score as SadTalker. However, when considering full scene coherence, our model significantly outperforms other models. This confirms our model's ability in scene context comprehension.

\subsection{Ablation Study}

\hspace{1em}\textbf{Ablation of lip references.} To check the effect of lip references $\mathcal{L}ip$ in the generation of face videos, we abort this latent condition during inference stage. This results in a degrading of about 0.3163 in SyncNet score, demonstrating the importance of latent condition $\mathcal{L}ip$ in guiding exact lip movements. We suppose the $\mathcal{L}ip$ can introduce more accurate temporally aligned conditions for lip-audio synchronization. 

\textbf{Adding additional facial expression references.} We also experiment on training the face generation model giving extra latent conditions like 3D Morphable Models\cite{blanz2023a} to guide facial expression generation. There resulting in no obvious perceptional improvement, but a slight increase in SyncNet score by 0.157. It is beneficial to add other explicit facial representations, but the landmarks has introduce most of information and the gain of adding more extra types of input becomes minor.

Due to space limitations, we put more ablations in the supplementary materials.

%% file: 5_conclusion.tex
\section{Conclusion}

\hspace{1em}In this paper, we introduce an interesting and novel setting for talking face video generation, which is taking the talking context into consideration when the context content such as audience or surroundings is provided. This setting is practical in applications such as face video inpainting/editing, multi-people interaction video generation, etc. 
In this paper, we focus on the video-inpainting-like scenario, and the specific target is to generate a talking face video driven by a given audio and a context video where the facial region is masked.
To achieve this target, we propose the TCCP, which is a two-stage generation pipeline, first transfer the implicit conditions in the context into explicit facial landmark sequence, and then generate the talking face videos. We also devise the MVControlNet, which is a multi-modal video generation diffusion model allowing for explicit control condition.   
In this paper, the generated content is limited to the head region, one may extend the region to whole body and generate the whole person using similar technique proposed, and we leave it for future work. Another limitation is only a single person result is provided in this main paper due to the data and computation resource limitation, even though our design, using the landmark as intermediate representation, holds a good potential for generalization. We believe this work has introduce a novel and practical setting to promote the talking face generation towards broader application, and also has provided solutions for the key problems introduced by the extra context condition. All those can be a valuable attempt, inspiring further works towards better solutions under the proposed setting. 

%% file: 6_supp.tex
\clearpage
\maketitlesupplementary
\renewcommand\thesection{\Alph{section}}
\setcounter{section}{0}

\section{Algorithm Details}

In this section, we introduce the implementation details of Architecture, Diffusion Process, Training Settings of Two-stage Cross-modal Control generation Pipeline in Table~\ref{tab:implementation}.

\begin{table}[t]
  \centering
  \resizebox{1\columnwidth}{!}{%
  \begin{tabular}{cc}
  \toprule[2pt]
    \multicolumn{2}{c}{\textbf{Architecture}} \\
    \hline
    Base channel & 128 \\
    Channel scale multiply & 1,2,4 \\
    Basic blocks per resolution & 2 + 1 down/up sample \\
    Downsample scale & 1,2,3 \\
    Self attention scale & 1,2,3 \\
    Latent based condition cross attetion scale & 1,2,3 \\
    Latent based condition embedding dimensions & 128 \\
    Attention head dimension & 64 \\
    Time step embedding dimensions & 128 \\
    Time step embedding MLP layers & 2 \\
    Latent diffusion auto-encoder downsampling factor & 16 \\
    \specialrule{0em}{1pt}{1.5pt}
    \midrule[2pt]
    \multicolumn{2}{c}{\textbf{Diffusion Process}} \\
    \hline
    Diffusion noise schedule & linear \\
    Diffusion steps & 1000 \\
    Prediction target & $\epsilon$ \\
    Sample method & DDPM \\
    \specialrule{0em}{1pt}{1.5pt}
    \midrule[2pt]
    \multicolumn{2}{c}{\textbf{Training Settings}} \\
    \hline
    Input/Output sequence length & 8 \\
    Input/Output shape & $8\times16\times16\times16$ \\
    Lip reference shape & $8\times1\times16\times16$ \\
    Video fps & 25 fps \\
    Audio Sample Rate & 16000 Hz \\
    Dropout & 0.1 \\
    Learning rate & 2e-4 \\
    Batch size & 192 \\
    Training steps & 150,000 \\
    Training hardware & 8 $\times$ NVIDIA 3090 Ti \\
    EMA & 0.9999 \\
    
  \specialrule{0em}{1pt}{1.5pt}
  \bottomrule[2pt]
  \end{tabular}%
  }
  \caption{The implementation details of TCCP}
  \label{tab:implementation}
  \end{table}

\section{Ablation Studies}

\begin{table}[t]
  \centering
  \resizebox{1\columnwidth}{!}{%
  \begin{tabular}{ccccc}
  \toprule[2pt]
    Model & {Sync. score $\uparrow$} & FID(Face) $\downarrow$ & FID(FullScene) $\downarrow$ & F.C. $\uparrow$ \\
    \hline
    One stage & 2.30 & 104.25 & 41.63 & 0.987 \\
    Two stage(gt lmk) & 2.99 & 68.05 & 26.42 & 0.988 \\
    Two stage(gen lmk) & 2.76 & 72.26 & 27.06 & 0.986 \\
\specialrule{0em}{1pt}{1.5pt}
\bottomrule[2pt]
\end{tabular}%
}
\caption{Comparison of quantitative results with one-stage method. F.C. denotes Frame Consistency. }
\label{tab:1 stage}
\end{table}

\begin{figure}
    \centering
    \includegraphics[width=1 \columnwidth]{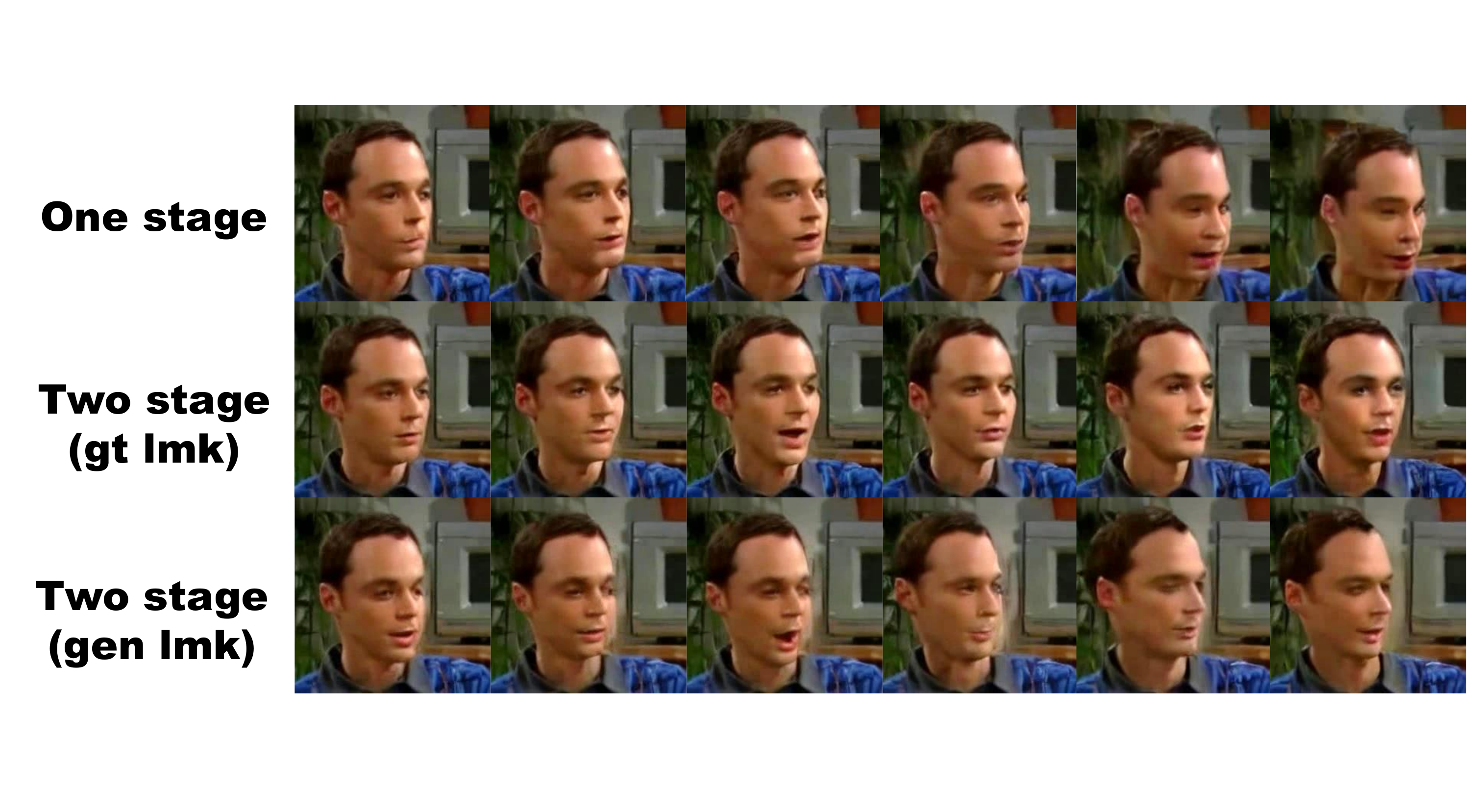}
    \caption{A comparison of generated face quality between one-stage and two-stage methods.}
    \label{fig:face_quality_1stage}
\end{figure}

\begin{figure*}
    \centering
    \includegraphics[width=1 \linewidth]{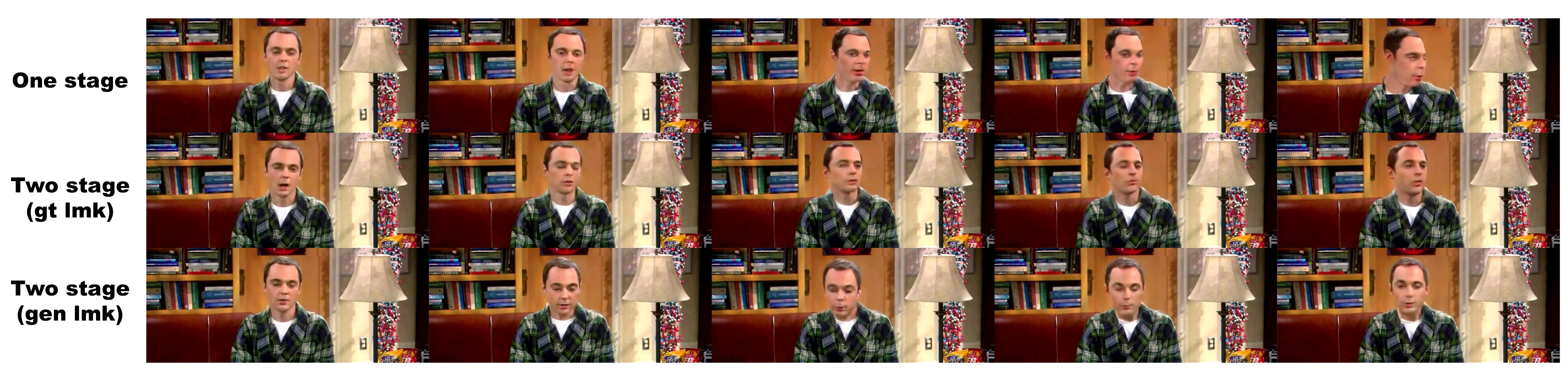}
    \caption{A comparison of generated full scene quality between one-stage and two-stage methods.}
    \label{fig:scene_quality_1stage}
\end{figure*}

\textbf{Comparison with one stage generation.}
We have demonstrated the effectiveness of our Two-stage Cross-modal Control generation Pipeline (TCCP) in Sec. 5.2 and Sec. 5.3. We further conduct an ablation experiment to explore the effectiveness of utilizing the facial landmarks as intermediate representation. The \textbf{one stage generation} skips taking head facial landmark $\mathcal{L}mk$ generation, and predicts  talking face video directly. One stage generation takes head-masked conversation scene video as video-based condition, and driving audio and lip sequence as latent condition. The model architecture and training strategy remain the same as TCCP, only removing intermediate representation $\mathcal{L}mk$. The one stage model is trained for 200k steps with batch size 192, and the quantitative results compared with TCCP is listed in Table~\ref{tab:1 stage}. Even trained with more steps, the one stage model degrades in audio-visual synchronization quality (Sync. score) and visual fidelity (FID). Without the help of facial landmarks, the one stage model has to learn the temporal correlation with driving audio and spatial coherent with context video simultaneously. This greatly increases training difficulty, resulting in relatively poor performance in both temporal and spatial coherence.

Figure~\ref{fig:face_quality_1stage} and Figure~\ref{fig:scene_quality_1stage} also provide qualitative comparisons between two strategies. As depicted in Figure~\ref{fig:face_quality_1stage}, the visual quality of faces generated by one stage model degrades severely with more iterations, displaying obvious distortion in the last two frames (after about 6 iterations).

Similarly, as shown in Figure~\ref{fig:scene_quality_1stage}, the full scene quality of one stage model also degrades with iterations. The head pose is getting strange, specially the shape of the character's neck, revealing a misunderstanding of the relation between the masked area and the full scene context.

These results demonstrates the effectiveness and importance of employing facial landmarks as an explicit and compact intermediate representation, to bridge the driving audio, video context and generated face videos. For dynamic display and comparison, please refer to our Supplementary Video.

\section{Implementation Details of User Study}

We sampled 10 talking contexts for user study, and the results of ControlNet, SadTalker and our TCCP were mixed and shuffled. 32 participants were provided with these 30 pairs of generated videos, each pair contains a generated face and the corresponding talking context with the generated face filled into the masked head area. The participants were shown the generated face videos first and asked to score the naturalness of face alone. After face scoring finished, the full conversation scenarios were given, and the participants were asked to judge the naturalness once again depending on the overall video quality. The videos used for user study and the questionaire are provided in Supplementary Material, in the form of powerpoint.